\documentclass[runningheads]{llncs}
\usepackage[T1]{fontenc}
\usepackage{cite}
\usepackage{graphicx}
\usepackage{amsmath}
\usepackage{amssymb,amsfonts}
\usepackage{textcomp}
\usepackage{xcolor}
\usepackage{soul}
\usepackage{subcaption}
\usepackage{footnote}

\usepackage{csquotes}
\usepackage{algorithm}
\usepackage{algpseudocode}
\usepackage{multirow}
\usepackage{multicol}
\usepackage{booktabs}
\usepackage{svg}
\usepackage{float}
\usepackage{hyperref}
\usepackage{xurl}      % allows better URL line breaks
\begin{document}
\title{Demographic-aware fine-grained visual recognition of pediatric wrist pathologies}

\titlerunning{Demographic-aware fine-grained recognition of pediatric wrist pathologies}
\author{Ammar Ahmed\inst{1}\orcidID{0009-0003-9984-4819} \and
Ali Shariq Imran\inst{1}\orcidID{0000-0002-2416-2878} \and
Zenun Kastrati\inst{2}\orcidID{0000-0002-0199-2377} \and 
Sher Muhammad Daudpota \inst{3}\orcidID{0000-0001-6684-751X}} 

\authorrunning{A. Ahmed et al.}
\institute{Department of Computer Science (IDI), Norwegian University of Science \& Technology (NTNU), Gjøvik, 2815, Norway \\
\email{ammaa@stud.ntnu.no, ali.imran@ntnu.no}\and
Department of Informatics, Linnaeus University, V\"axj\"o, 351 95, Sweden \\ 
\email{zenun.kastrati@lnu.se}\and
Department of Computer Science, Sukkur IBA University, Sukkur, 65200, Pakistan \\ \email{sher@iba-suk.edu.pk}}
\maketitle              % typeset the header of the contribution
\begin{abstract}
Pediatric wrist pathologies recognition from radiographs is challenging because normal anatomy changes rapidly with development: evolving carpal ossification and open physes can resemble pathology, and maturation timing differs by sex. Image-only models trained on limited medical datasets therefore risk confusing normal developmental variation with true pathologies. We address this by framing pediatric wrist diagnosis as a fine-grained visual recognition (FGVR) problem and proposing a demographic-aware hybrid convolution--transformer model that fuses X-rays with patient age and sex. To leverage demographic context while avoiding shortcut reliance, we introduce progressive metadata masking during training. We evaluate on a curated dataset that mirrors the typical constraints in real-world medical studies. The hybrid FGVR backbone outperforms traditional and modern CNNs, and demographic fusion yields additional gains. Finally, we show that initializing from a fine-grained pretraining source improves transfer relative to standard ImageNet initialization, suggesting that label granularity, even from non-medical data, can be a key driver of generalization for subtle radiographic findings.  

\keywords{Pediatric fractures \and fracture detection \and wrist pathologies \and fine-grained classification \and Multimodal fusion}
\end{abstract}

\section{Introduction}
Wrist fractures represent up to 75\% of hand injuries in emergency departments and pose a significant risk of long-term impairment if misdiagnosed \cite{hoynak2022wrist}. Given the wrist’s complex anatomy and developmental variability, accurate radiographic interpretation is essential. Recent advances in computer vision have demonstrated promise for automated abnormality detection in musculoskeletal imaging \cite{adams_robert_yi_babyn_2020}. 
However, such methods typically rely on large, labeled datasets, which are scarce in medical domains, and often process X-rays without contextual patient information.

To overcome the limitations of image-only learning, related work in multimodal frameworks that fuse patient metadata with imaging have been successfully applied in dermatology and oncology domains. Studies have shown consistent accuracy improvements by incorporating age, sex, and anatomical site information
% alongside dermoscopic or histopathology 
images 
\cite{ningrum2021deep,nunnari2020study,cai2023multimodal,thomas2021combining}. 
For example, Thomas et al. \cite{thomas2021combining} fused demographic variables with dermoscopy images, while Cai et al. \cite{cai2023multimodal} applied ViTs with metadata encoders, achieving 93.8\% melanoma classification accuracy. Recently, Hinterwimmer et al. \cite{florian2024} extended this approach to musculoskeletal imaging, reporting gains in bone tumour classification by integrating X-rays and metadata. Despite these advances, metadata integration has not been systematically investigated for wrist pathology recognition.

Pediatric wrist radiographs are particularly challenging because anatomy changes rapidly with development: open physes and progressive carpal ossification can mimic pathology, and maturation differs by sex \cite{bian_hand_wrist_bone_age_2020,kox_amphys_protocol_2018,roche_skeletal_maturity_1974}. This yields small inter-class differences (subtle fractures vs.\ normal irregularities) and large intra-class variation, closely matching the conditions of fine-grained visual recognition (FGVR) \cite{zhao_fine_grained_segmentation_2017,ahmed2024learning}. We therefore study pediatric wrist recognition under three hypotheses:
\textbf{H1:} FGVR-style hybrid convolution--transformer architectures outperform purely convolutional baselines, including modern CNNs such as ConvNeXt \cite{woo2023convnextv2} (and detection-based baselines such as YOLO \cite{Jocher_Ultralytics_YOLO_2023}), under matched training settings.
\textbf{H2:} Fusing age and sex improves accuracy and robustness by providing developmental context.
\textbf{H3:} Fine-grained pretraining transfers better than ImageNet pretraining because it encourages representations sensitive to subtle visual differences \cite{cui_large_scale_fine_grained_2018}.
% \subsection{Contributions}
Our contributions are as follows:
\begin{enumerate}
    \item \textbf{FGVR inductive bias for pediatric wrists:} We argue and empirically test that pediatric wrist abnormality recognition is well-modeled as an FGVR problem under strong developmental variation.
    \item \textbf{Demographic-aware multimodal model:} We use a fusion strategy that integrates age and sex while discouraging shortcut learning via progressive metadata masking.
    \item \textbf{Fine-grained pretraining:} We show that fine-grained non-medical pretraining improves transfer relative to ImageNet initialization.
    % \item \textbf{Comprehensive evaluation:} We validate on a curated three-class subset and a larger fracture-vs-no fracture cohort derived from GRAZPEDWRI-DX.
\end{enumerate}

% \section{Related Work}
% \section{Related Work}
% \paragraph{Multimodal learning with clinical metadata.}
% A growing body of work shows that fusing medical images with structured patient information can improve diagnostic performance, particularly when image evidence is ambiguous or when clinically meaningful confounders exist. In dermatology, multiple studies report accuracy gains by combining dermoscopy images with demographics and anatomical context such as age, sex, and lesion site \cite{ningrum2021deep,nunnari2020study,thomas2021combining}. More recently, transformer-based designs have incorporated dedicated metadata encoders to learn joint representations from images and tabular variables \cite{cai2023multimodal}. Beyond dermatology, multimodal fusion has been explored in other clinical domains, including musculoskeletal imaging; Hinterwimmer et al.\ \cite{florian2024} demonstrated improved bone tumour classification by integrating radiographs with patient metadata. 

% \paragraph{Gap in pediatric wrist pathology recognition.}
% Despite these advances, demographic-aware multimodal learning has not been systematically studied for \emph{pediatric wrist pathology recognition}, where age- and sex-dependent skeletal maturation substantially modulates normal radiographic appearance. This motivates our demographic-aware FGVR framework, which explicitly leverages age and sex while regularizing their use to reduce shortcut reliance.

\section{Methods}
\subsection{Dataset Curation \& Patient Metadata}
We curated two classification datasets derived from GRAZPEDWRI-DX dataset \cite{nagy2022pediatric}. The dataset comprises 20,327 wrist radiographs from 6,091 pediatric patients with a mean age of 10.9 years (range 0.2–19 years)
% , collected at the Division of Paediatric Radiology, Department of Radiology, Medical University of Graz, Austria.
with 2,688 females, 3,402 males, and one patient with unknown sex. The images include both posteroanterior and lateral projections. GRAZPEDWRI-DX spans nearly the full pediatric age range from infancy to late adolescence. This wide coverage captures multiple stages of skeletal development, including progressive carpal ossification and age-dependent changes in the distal radius/ulna physis.

% Additionally, the availability of age and sex enables explicit conditioning on developmental context, which is critical for separating normal maturation-related appearance changes from true abnormalities

% We curated two specialized datasets from the GRAZPEDWRI-DX dataset \cite{nagy2022pediatric}. For the first ``limited'' dataset, we use the same curation, splitting, and augmentation process as in our previous works \cite{ahmed2024learning, 10648070}. In this study, however, we exclude the ``metal'' class, as our focus is on inherent wrist pathologies rather than the presence of external objects visible in the radiographs. The number of instances in the final split for the limited set (Three Wrist Pathologies) is shown in the left sub-table of Table \ref{tab:combined_split}. In addition, we also curated an additional dataset containing all images from the dataset. To achieve this, we categorized the dataset into two classes: \enquote{Fracture} and \enquote{No Fracture}. This would then serve to further evaluate the influence of metadata integration when utilizing the entire dataset. The distribution of training, validation, and test sets for this dataset is also detailed in Table \ref{tab:combined_split}. In this case, the \enquote{No Fracture} class was augmented to approximate the number of instances in the \enquote{fracture} class. The final split ratio for this specific set is 86.65\% for training, 10.68\% for validation, and 2.66\% for testing. 

% We curated two datasets from GRAZPEDWRI-DX \cite{nagy2022pediatric}.  
% (1) 

\begin{table}
\centering
\caption{Instances in training, validation, and test sets from the curated datasets.}
\setlength{\tabcolsep}{4pt} % tighter columns (optional)
\small % or \footnotesize

\begin{tabular}{l c c c}
\hline
\multicolumn{4}{c}{\textbf{Three Wrist Pathologies (Limited set)}} \\
\hline
Class & Training & Validation & Test \\
\hline
Boneanomaly  & 392 & 98  & 119 \\
Fracture     & 400 & 100 & 120 \\
Soft-tissue  & 376 & 94  & 115 \\
\hline
Total        & 1168 & 292 & 354 \\
\midrule
\multicolumn{4}{c}{\textbf{Fracture vs. No Fracture (Entire set)}} \\
\midrule
Class & Training & Validation & Test \\
\hline
Fracture     & 11854 & 1357 & 338 \\
No Fracture  & 10164 & 1357 & 338 \\
\hline
Total        & 22018 & 2714 & 676 \\
\hline
\end{tabular}

\label{tab:combined_split}
\end{table}

\begin{figure}
\centering
\includegraphics[width=0.8\linewidth]{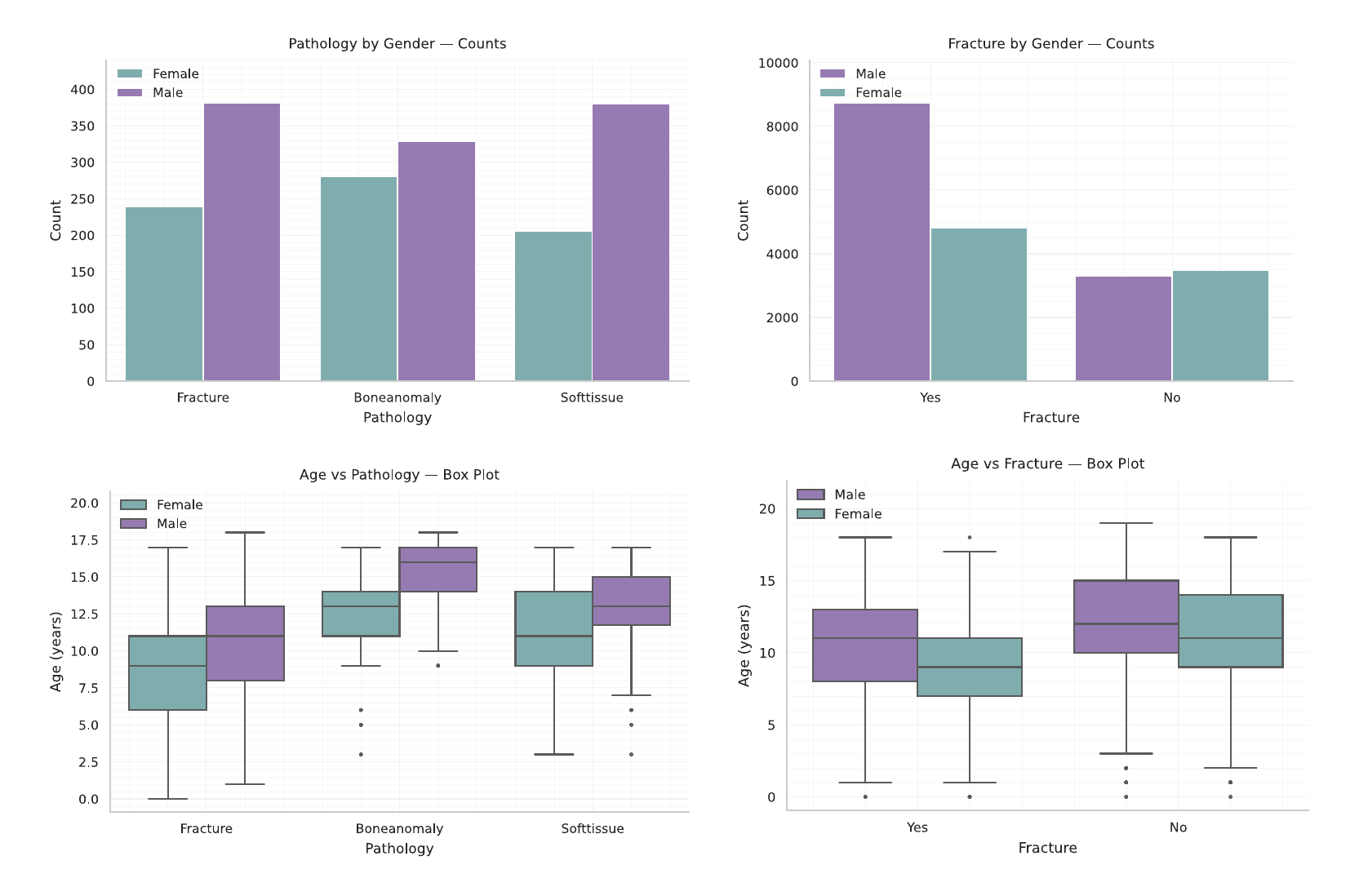}
\caption{Bar and box plots for gender and age, respectively, for both curated datasets.}
\label{fig1:countp}
\end{figure}

The first curated dataset is a limited three-class pathology subset comprising fracture, bone anomaly, and soft-tissue abnormality. All other labels were excluded because, after restricting the instances with exactly one pathology label per image (to avoid multi-label ambiguity), the remaining classes contained too few examples for reliable training. The second curated dataset is a binary “Fracture vs. No Fracture” that uses the full set of available images. Final dataset splits are reported in Table~\ref{tab:combined_split}. Figure~\ref{fig1:countp} summarizes demographic distributions across pathologies.

\subsection{Preprocessing and Augmentation}
All radiographs are resized to $224\times224$. To improve robustness to acquisition variability, we apply training-time augmentation with Keras \texttt{ImageDataGenerator}: rotations ($\pm15^\circ$), width/height shifts (0.2/0.1), shear (0.2), zoom (0.1), and brightness jitter $(0.7,1.3)$. We additionally use ZCA whitening and constant padding for empty regions (\texttt{fill\_mode="constant"}).
\begin{align*}
& \texttt{rotation\_range}=15, \\
& \texttt{width\_shift\_range}=0.2, \\
& \texttt{height\_shift\_range}=0.1, \\
& \texttt{horizontal\_flip}=\texttt{True}, \\
& \texttt{shear\_range}=0.2, \\
& \texttt{brightness\_range}=(0.7, 1.3), \\
& \texttt{zoom\_range}=0.1, \\
& \texttt{fill\_mode}=\texttt{"constant"}, \\
& \texttt{zca\_whitening}=\texttt{True}.
\end{align*}

% While the selection of augmentation magnitudes is inevitably heuristic, 
We chose these values to provide effective regularization without introducing unrealistic transformations that could move relevant anatomy out of frame or substantially distort exposure.
% In particular, we avoid excessive rotations and shifts and restrict brightness and zoom to ranges that preserve clinically meaningful structure.

\subsection{Architectural Details}

\begin{figure*}
\centering
\includegraphics[width=1.01\linewidth]{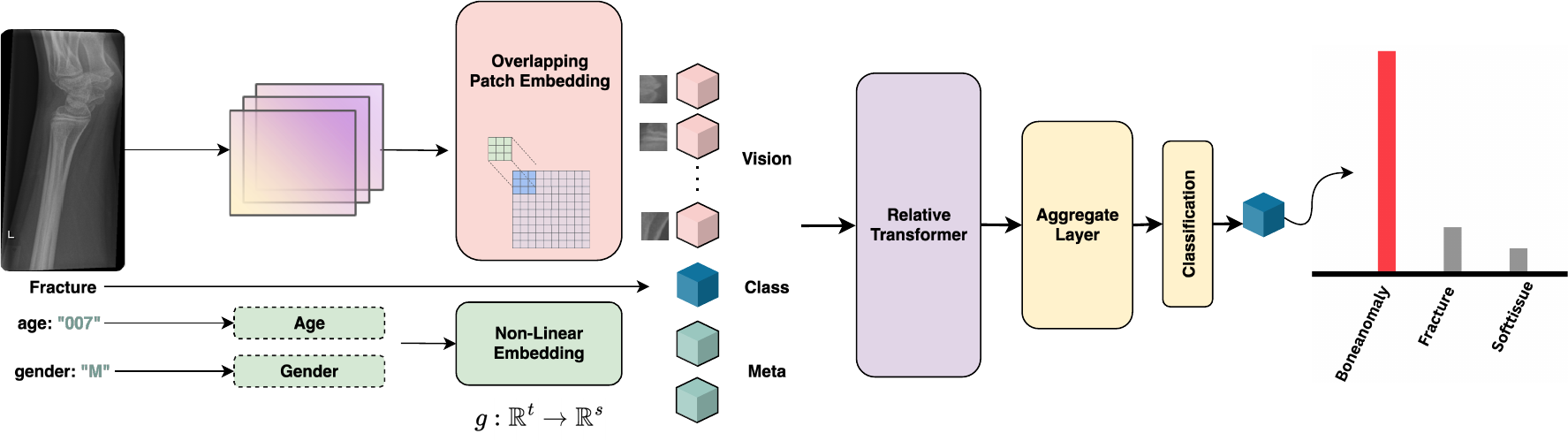}
\caption{The architecture pipeline integrates visual, class, and meta tokens within a transformer framework.}
\label{meta-arch}
\end{figure*}

This work utilizes a hybrid architecture that integrates both visual data and meta-information, drawing inspiration from the base MetaFormer architecture \cite{diao2022metaformer}. As depicted in Fig. \ref{meta-arch}, the architecture is organized into five stages. The first stage (S0) starts with a basic 3-layer convolutional stem, which processes raw input images. Subsequent stages progressively downsample the feature maps and increase channel dimensionality. Stages S1 and S2 consist of MBConv blocks enhanced with squeeze-and-excitation mechanism. The final two stages, S3 and S4, transition to transformer-based blocks that employ relative positional encoding. This design allows the model to capture global
dependencies, which are crucial for subtle pathological differences that span larger contexts (e.g., comparing radius vs.
ulna contours or detecting diffuse soft tissue swelling).

Downsampling is applied at the beginning of each stage. In S3 and S4, overlapping patch embedding is implemented using convolutions with a stride of 2 and zero-padding, effectively reducing resolution while maintaining contextual overlap. Table \ref{tab:metaformer-series} outlines the configuration details of the three variants of the architecture used in our experiments.

% \begin{figure*}
% \centering
% \includegraphics[width=0.7\textwidth]{Figures/MetaFormer_Architecture.png}
% \caption{The architecture integrates visual, class, and meta tokens within a transformer framework. Meta-information is embedded using a non-linear projection. Fusion occurs through a Relative Transformer Layer, and classification is performed using the final class token.}
% \label{meta-arch}
% \end{figure*}

\begin{table}
\centering
\caption{Configuration of the architecture stages employed in this study. B: number of blocks, and H: hidden dimensions for each stage.}
\resizebox{0.5\columnwidth}{!}{%
    \begin{tabular}{l c c c}
    \hline
    Stage & FG-0 & FG-1 & FG-2\\
    \hline 
        S0 & B=3 H=64 & B=3 H=64 & B=3 H=128\\
        S1 & B=2 H=96 & B=2 H=96 & B=2 H=128\\
        S2 & B=3 H=192 & B=6 H=192 & B=6 H=256\\
        S3 & B=5 H=384 & B=14 H=384 & B=14 H=512\\
        S4 & B=2 H=768 & B=3 H=768 & B=3 H=1024\\
    \hline                                                      
  \end{tabular}}
  \label{tab:metaformer-series}
\end{table}

To overcome the lack of spatial awareness in standard self-attention, the model incorporates a relative position bias. Specifically, a bias matrix \( B \in \mathbb{R}^{((M^2 + N) \times (M^2 + N))} \) is added during attention computation:
\begin{equation}
\textit{Attention}(Q, K, V) = \text{Softmax}\left( \frac{QK^T}{\sqrt{d}} + B \right) V.
\end{equation}

Here, $Q$, $K$, and $V$ are the query, key, and value matrices of size \( \mathbb{R}^{((M^2 + N) \times d)} \), where $M^2$ corresponds to the number of image patches and $N$ is the count of extra tokens (e.g., class and meta tokens). The bias matrix $\hat{B}$ is parameterized, supporting relative positions from $-M-1$ to $M+1$. A shared bias is applied for non-visual tokens since their spatial position is undefined.
% \begin{figure}
% \centering
% \includesvg[width=\linewidth]{Figures/MetaFormer copy 2.drawio}
% \caption{The architecture pipeline integrates visual, class, and meta tokens within a transformer framework.}
% \label{meta-arch}
% \end{figure}
The Relative Transformer Block comprises multi-head self-attention (MSA), layer normalization (LN), and a multi-layer perceptron (MLP), applied to the input token sequence $z_0$:
\begin{equation}\label{eq:combined_equation}
\begin{aligned}
z_0 &= [x_{\text{class}} ; x^1_{\text{meta}} , \ldots, x^{n-1}_{\text{meta}} ; x^1_{\text{vision}}, \ldots, x^m_{\text{vision}}], \\
z'_i &= \text{MSA}(\text{LN}(z_{i-1})) + z_i, \\
z_i &= \text{MLP}(\text{LN}(z'_i)) + z'_i. 
\end{aligned}
\end{equation}

In this formulation, $z_i \in \mathbb{R}^{(M^2 + N) \times d}$ captures the updated token sequence at each transformer layer. Stages S3 and S4 each generate a class token, $z^{1}_{\text{class}}$ and $z^{2}_{\text{class}}$, which encode combined visual and meta representations. Since these tokens may differ in dimension, $z^{1}_{\text{class}}$ is first projected to match the shape of $z^{2}_{\text{class}}$ using an MLP. The two are then concatenated and passed through a 1D convolution followed by normalization:
\begin{equation}\label{eq:combined_equation_2}
\begin{aligned}
\hat{z}^{\text{1}}_{\text{class}} &= \text{MLP}(\text{LN}(z^{\text{1}}_{\text{class}})), \\
z_{\text{class}} &= \text{Conv1d}(\text{Concat}(\hat{z}^{\text{1}}_{\text{class}}, z^{\text{2}}_{\text{class}})), \\
y &= \text{LN}(z_{\text{class}}).
\end{aligned}
\end{equation}

The final output $y$ aggregates information across multiple scales and token types. When using external attributes as metadata, these are encoded as vectors and embedded using a fully connected non-linear function \( f : \mathbb{R}^n \rightarrow \mathbb{R}^d \). 
% In our setup, the demographic attributes consist of age and gender. The age attribute was normalized using min–max scaling, while the gender attribute was one-hot encoded. This results in a 2D input vector, which is then projected into the same embedding space as the vision tokens.
% , ensuring consistency between visual and meta representations.
% For instance, our curated dataset with two attribute values would produce a 2D input vector, which is then projected into the same embedding space as the vision tokens.
% To maintain the model’s ability to learn important visual features, especially when metadata is less informative or potentially overfitting, a regularization strategy is applied during training that progressively masks parts of the meta-information. This balances the reliance on both vision and auxiliary data.
To prevent overreliance on metadata (and potential overfitting), we apply progressive metadata masking during training, forcing the model to learn robust visual features while still leveraging auxiliary signals.

\subsection{Experimental Settings}
% A range of well-established models has been incorporated in our study alongside our fine-grained metadata-aware approach for comparison. All these baseline models were pre-trained on ImageNet \cite{deng2009imagenet}. These include ViT \cite{dosovitskiy2020image}, VGG16 \cite{simonyan2015very}, GoogleNet \cite{szegedy2015going}, EfficientNetV2 \cite{tan2021efficientnetv2}, AlexNet \cite{krizhevsky2012imagenet}, DenseNet201 \cite{huang2018densely}, InceptionV3 \cite{szegedy2014going}, and ResNet50 \cite{he2015deep}. 

We compare against traditional and modern CNNs pre-trained on ImageNet \cite{deng2009imagenet,dosovitskiy2020image,tan2021efficientnetv2,krizhevsky2012imagenet,huang2018densely,szegedy2015going,he2015deep,woo2023convnextv2,liu2021swintransformer,Jocher_Ultralytics_YOLO_2023}. All models underwent training for 100 epochs using a resolution of $224\times224$ and a batch size of 32. Adam optimizer with $lr=10^{-3}$ was the default for baselines, AdamW with base $lr=10^{-4}$ for MetaFormer, and SGD with base $lr=10^{-2}$ for YOLO. We quantify uncertainty using 95\% bootstrap confidence intervals obtained by resampling test examples with replacement (percentile method, $B=2000$). Code and the curated datasets are publicly available at \url{https://github.com/ammarlodhi255/demographic-aware-fgvr-wrist-pathology-recognition.git}.

% , warmup learning rate \(\alpha = 5 \times 10^{-7}\), minimum learning rate was \(\eta_m = 1 \times 10^{-5}\), and weight decay was \(\wp = 5 \times 10^{-2}\)
% 

\section{Results \& Discussion}

% \begin{table}
% \caption{Performance evaluation of different baseline neural networks and the MetaFormer on only image data.}
% \centering
%     \begin{tabular}{l c}
%     \hline
%     Model & Test Accuracy (Vision Only)\\
%     \hline 
%       ViT & 62.71\% \\
%       VGG16 & 63.28\% \\
%       GoogleNet & 70.34\% \\
%       EfficientNetV2 & 72.32\% \\
%       AlexNet & 74.29\% \\
%       DenseNet201 & 76.84\% \\
%       InceptionV3 & 77.44\% \\
%       ResNet50 & 77.68\% \\
%       MetaFormer FGVR (Base) & \textbf{79.40\%} \\
%     \hline                                                      
%   \end{tabular}
%   \label{tab:baseline_image_only}
% \end{table}

% Preamble (once):
% \usepackage{booktabs}

% \begin{table}
% \centering
% \caption{Accuracy (\%) of different architecture configurations on vision, vision + meta through early fusion, and late fusion.}
%     \begin{tabular}{l c c c}
%     \hline
%     Config & Vision & Early Fusion & Late Fusion\\
%     \hline 
%         FG-0 & 79.4\% & 79.1\% & 80.5\%\\
%         FG-1 & 78.5\% & 80.5\% & 79.1\%\\
%         FG-2 & 77.4\% & 78.5\% & 79.4\%\\
%         \textbf{FG-2-inat (Ours)} & \textbf{79.9\%} & \textbf{81.1\%} & \textbf{81.4\%}\\
%     \hline                                                      
%   \end{tabular}
%   \label{tab:metaformer-config}
% \end{table}

\subsection{Baseline Comparison}
Table~\ref{tab:baseline_image_only} shows that baseline MetaFormer (vision only) outperforms widely used CNNs and more recent architectures (Swin Transformer, YOLO11s, and Conv\allowbreak NeXtV2-Tiny). This gives credence to the fact that transformer-based architectures with fine-grained design are better suited to capture subtle morphological cues in pediatric wrist radiographs. At the bottom, our final fusion variant yields the highest accuracy. 
% To quantify uncertainty without repeated retraining, we report 95\% confidence intervals computed via percentile bootstrap on the test set (B=2000 resamples). At the bottom, we have our proposed fusion variants, showing that integrating metadata improves performance.

% MetaFormer (vision-only) outperforms both classic CNNs and recent backbones (Swin, YOLO11s, ConvNeXtV2-Tiny) as shown in Table~\ref{tab:baseline_image_only}, suggesting fine-grained Transformer designs better capture subtle pediatric wrist cues. To quantify uncertainty, we report accuracy with 95\% percentile-bootstrap CIs (B=2000). Our fusion variants further improve performance via demographic metadata integration.

% MetaFormer (vision-only) achieves the highest test accuracy among the compared image-only baselines (Table~\ref{tab:baseline_image_only}).  

\begin{table}
  \centering
  \caption{Vision-only baselines on the test set. MetaFormer fusion variant (bottom) achieve the best performance.}
  \label{tab:vision_only_baselines_compact}
  \resizebox{0.65\linewidth}{!}{%
  \begin{tabular}{l c}
    \toprule
    \textbf{Model} & \textbf{Accuracy (\%) (95\% CI)} \\
    \midrule
    ViT                & \(59.0\ (54.0\text{--}64.1)\) \\
    AlexNet            & \(66.1\ (61.3\text{--}71.2)\) \\
    YOLO11s            & \(69.5\ (64.7\text{--}74.3)\) \\
    VGG16              & \(70.1\ (65.5\text{--}74.9)\) \\
    GoogleNet          & \(72.9\ (68.1\text{--}77.4)\) \\
    ResNet50           & \(74.0\ (69.8\text{--}78.5)\) \\
    DenseNet201        & \(74.9\ (70.3\text{--}79.4)\) \\
    ConvNeXtV2    & \(76.0\ (71.5\text{--}80.2)\) \\
    InceptionV3        & \(77.1\ (72.6\text{--}81.4)\) \\
    Swin Transformer   & \(77.4\ (72.9\text{--}81.6)\) \\
    EfficientNetV2     & \(78.0\ (73.7\text{--}81.9)\) \\
    
    \textbf{MetaFormer (Base)} & \(\mathbf{79.1}\ (\boldsymbol{74.9\text{--}83.3})\) \\
    \midrule
    \textbf{MetaFormer-FG-2-inat-vision (Ours)} & \(\mathbf{79.9}\ (\mathbf{75.9}\textbf{--}\mathbf{83.9})\) \\
    \textbf{MetaFormer-FG-2-inat-fusion (Ours)} & \(\mathbf{82.2}\ (\mathbf{78.2}\textbf{--}\mathbf{86.2})\) \\

    % \textbf{MetaFormer-FG-2-inat-late (Ours)} & \(\mathbf{83.6}\ (\mathbf{79.7}\textbf{--}\mathbf{87.3})\) \\
    \bottomrule
  \end{tabular}%
  }
  \label{tab:baseline_image_only}
\end{table}

% \subsection{Metadata Integration \& Pretraining}

% \subsection{Ablation Analysis}
% Integrating demographic metadata (age, sex) consistently improved performance (Table~\ref{tab:metaformer-config}). Both early and late fusion surpassed vision-only models, with late fusion achieving the highest gains in two out of three configurations. This finding aligns with the intuition that age and sex provide essential developmental context, particularly relevant in pediatrics, where normal ossification patterns vary across individuals. Late fusion may be more robust to noise in individual modalities, as combining information at a later stage allows the final decision to benefit from multiple sources, potentially mitigating the impact of noise or errors in any single modality.

% Attribute-level analysis (Table~\ref{tab:indiv_analysis}) further indicates that both age and sex independently contribute to performance, but their combination yields the largest benefit, supporting the clinical relevance of demographic context.

% We then selected the ``FG-2'' configuration and used the weights obtained from training it on the iNaturalist dataset \cite{vanhorn2018inaturalist} to create our ``FG-2-inat'' configuration. This surpassed all other configurations, whether using image data alone or in conjunction with metadata. This configuration reached 81.4\%, a 1.5\% improvement over image-only training. It seems that the fine-grained pretraining improved generalization beyond ImageNet pretraining, reflecting the value of transferring from domains where subtle inter-class variations dominate. 
\subsection{Ablation Analysis}
Integrating demographic metadata improves performance across configurations (Table~\ref{tab:metaformer-config}). In particular, \textit{token fusion} consistently outperforms \textit{vision} and \textit{early fusion}, suggesting that injecting demographic information at the representation/token level is more effective than concatenating it at the input. 
% This aligns with the clinical intuition that age and sex provide developmental context that helps disentangle normal maturation-related anatomy from pathology, especially in pediatrics where ossification patterns vary substantially across individuals.
Attribute-level analysis (Table~\ref{tab:indiv_analysis}) further indicates that both age and sex contribute independently to performance, while combining them yields the largest improvement, supporting the clinical relevance of demographic context.

We selected the \enquote{FG-2} configuration and initialized it with weights pretrained on iNaturalist \cite{vanhorn2018inaturalist}, yielding \enquote{FG-2-inat}. This model achieves the strongest overall performance, surpassing all other configurations under both fusion strategies. Compared to ImageNet initialization, fine-grained pretraining provides a clear advantage, suggesting that supervision from fine-grained domains, where subtle inter-class differences dominate, induces representations that transfer well to pediatric wrist abnormalities.

% \begin{table}
% \centering
% \caption{Joint evaluation of architecture configuration and fusion strategy reported with test accuracy (\%) (95\% CI).}
% \resizebox{\columnwidth}{!}{%
% \begin{tabular}{l c c c c}
% \hline
% \textbf{Config} & \textbf{Vision} & \textbf{Late Fusion} & \textbf{Early Fusion} & \textbf{Token Fusion}\\
% \hline 
% FG-0 & \({79.1}\ ({74.9\text{--}83.3})\) & \(79.4\ (75.1\text{--}83.3)\) & \(75.1\ (70.6\text{--}79.7)\) & \(78.5\ (74.0\text{--}82.5)\) \\
% FG-1 & \(77.4\ (73.2\text{--}81.6)\) & \(78.0\ (73.4\text{--}82.2)\) & \(\mathbf{76.0}\ (\mathbf{71.5}\textbf{--}\mathbf{80.2})\) & \(79.7\ (75.4\text{--}83.6)\) \\
% FG-2 & \(78.5\ (74.3\text{--}82.5)\) & \(80.5\ (76.3\text{--}84.5)\) & \(74.9\ (70.3\text{--}79.4)\) & \(73.4\ (68.6\text{--}78.0)\) \\
% \textbf{FG-2-inat (Ours)} & \(\mathbf{82.2}\ (\mathbf{78.2}\textbf{--}\mathbf{86.2})\) & \(\mathbf{83.6}\ (\mathbf{79.7}\textbf{--}\mathbf{87.3})\) & \(75.1\ (70.6\text{--}79.7)\) & \(\mathbf{82.2}\ (\mathbf{78.2}\textbf{--}\mathbf{86.2})\) \\
% \hline
% \end{tabular}%
% }
% \label{tab:metaformer-config}
% \end{table}

\begin{table}
\centering
\caption{Joint evaluation of architecture configuration and fusion strategy reported with test accuracy (\%) (95\% CI).}
\setlength{\tabcolsep}{10pt}
\renewcommand{\arraystretch}{1.15}
\resizebox{0.82\columnwidth}{!}{%
\begin{tabular}{l c c c}
\hline
\textbf{Configuration} & \textbf{Vision} & \textbf{Early Fusion} & \textbf{Token Fusion (Ours)}\\
\hline 
FG-0 & \({79.1}\ ({74.9\text{--}83.3})\) & \(75.1\ (70.6\text{--}79.7)\) & \(78.5\ (74.0\text{--}82.5)\) \\
FG-1 & \(77.4\ (73.2\text{--}81.6)\) & \(\mathbf{76.0}\ (\mathbf{71.5}\textbf{--}\mathbf{80.2})\) & \(79.7\ (75.4\text{--}83.6)\) \\
FG-2 & \(78.5\ (74.3\text{--}82.5)\) & \(74.9\ (70.3\text{--}79.4)\) & \(73.4\ (68.6\text{--}78.0)\) \\
\textbf{FG-2-inat (Ours)} & \(\mathbf{79.9}\ (\mathbf{75.9}\textbf{--}\mathbf{83.9})\) & \(75.1\ (70.6\text{--}79.7)\) & \(\mathbf{82.2}\ (\mathbf{78.2}\textbf{--}\mathbf{86.2})\) \\
\hline
\end{tabular}%
}
\label{tab:metaformer-config}
\end{table}

Table \ref{tab:masking_ablation} shows that if masking is too weak, the network can overfit to metadata-correlated shortcuts (reducing robustness when demographic priors shift). Conversely, overly strong or constant masking suppresses informative auxiliary signals and prevents the model from learning a stable joint representation. Our \textit{progressive} masking schedule provides the best trade-off, yielding the highest test accuracy by gradually reducing metadata availability over training.

\begin{table}[H]
\centering
\caption{Accuracies obtained by FG-2-inat-default configuration on metadata attributes.}
\resizebox{0.65\columnwidth}{!}{%
    \begin{tabular}{l c}
    \hline
    \textbf{Attribute} & \textbf{Accuracy (\%) (95\% CI)} \\
    \hline 
        Vision + Age & \(79.7\ (75.4\text{--}83.6)\) \\
        Vision + Gender & \(79.1\ (74.9\text{--}83.3)\) \\
        Vision + Age + Gender & \(\mathbf{82.2}\ (\mathbf{78.2}\text{--}\mathbf{86.2})\)\\
    \hline                                                      
  \end{tabular}}
  \label{tab:indiv_analysis}
\end{table}

\begin{table}
\centering
\caption{Effect of masking schedule and probability on test accuracy.}
\setlength{\tabcolsep}{10pt}
\renewcommand{\arraystretch}{1.15}
\resizebox{0.7\columnwidth}{!}{%
\begin{tabular}{l c c}
\hline
\textbf{Mask Type} & \textbf{Mask Prob} & \textbf{Accuracy (\%) (95\% CI)} \\
\hline
\multirow{4}{*}{Constant}
  % & $0.05$ & $81.0 \pm 0.01$ \\
  & $0.1$ & \(79.7\ (75.4\text{--}83.9)\) \\
  & $0.2$ & \(79.1\ (75.1\text{--}83.3)\) \\
  & $0.3$ & \(79.7\ (75.1\text{--}83.6)\) \\
  & $0.5$ & \(79.4\ (75.1\text{--}83.3)\) \\
\hline
\multirow{4}{*}{Linear}
  % & $0.05$ & $80.9 \pm 0.01$ \\
  & $0.1$ & \(79.7\ (75.4\text{--}83.6)\) \\
  & $0.2$ & \(78.8\ (74.6\text{--}83.1)\) \\
  & $0.3$ & \(78.8\ (74.6\text{--}83.1)\) \\
  & $0.5$ & \(79.4\ (75.1\text{--}83.6)\) \\
  \midrule
  \textbf{Progressive (Ours)} & Scheduled & \(\mathbf{82.2}\ (\mathbf{78.2}\textbf{--}\mathbf{86.2})\) \\
\hline
\end{tabular}}
\label{tab:masking_ablation}
\end{table}

\subsection{Interpretability with XAI}
\begin{figure}
\centering
\includegraphics[width=0.8\linewidth]{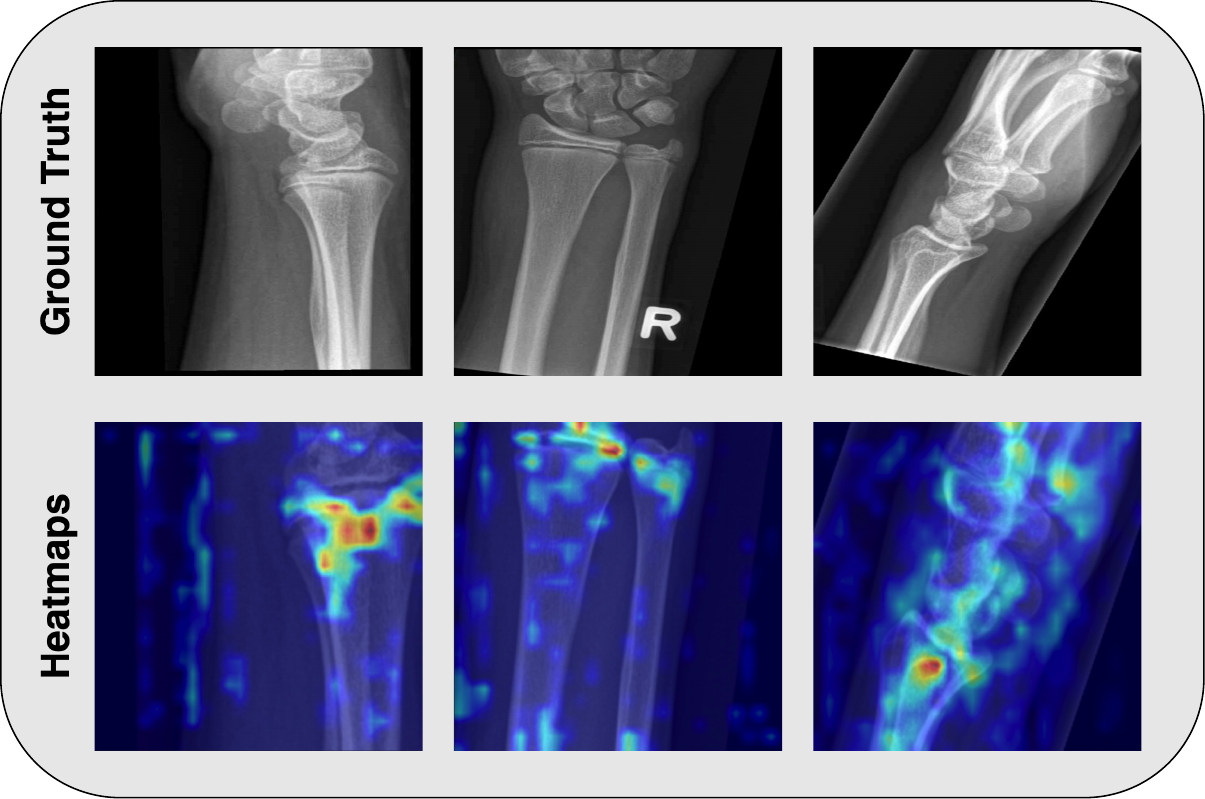}
\caption{Heatmaps from the final convolutional layer in stage 2 (S2) of our configuration ``FG-2-inat'' on a limited dataset.}
\label{fig1:heatmaps}
\end{figure}

% Grad-CAM visualizations (Fig.\ref{fig1:heatmaps}) show that MetaFormer’s high-saliency regions often overlap with clinically relevant wrist anatomy, and in several cases concentrate near the suspected fracture site or abnormal osseous structures. However, some maps are spatially diffuse, indicating that the model’s evidence is not always well localized and may occasionally include non-pathological cues. This suggests that metadata fusion improves classification performance, but interpretability and localization could be further strengthened (e.g., via attention/region constraints or localization-aware training).
% As shown in Fig.\ref{fig:roc}, incorporating metadata improves fracture discrimination while maintaining comparable performance for bone anomaly and soft-tissue recognition.
Grad-CAM's output (Fig.\ref{fig1:heatmaps}) show MetaFormer’s salient regions often overlap clinically relevant wrist anatomy (e.g., near suspected fractures or abnormal bone), though some maps are diffuse, suggesting imperfect localization and occasional reliance on non-pathological cues. 

  \begin{figure}
\centering
\includegraphics[width=0.9\linewidth]{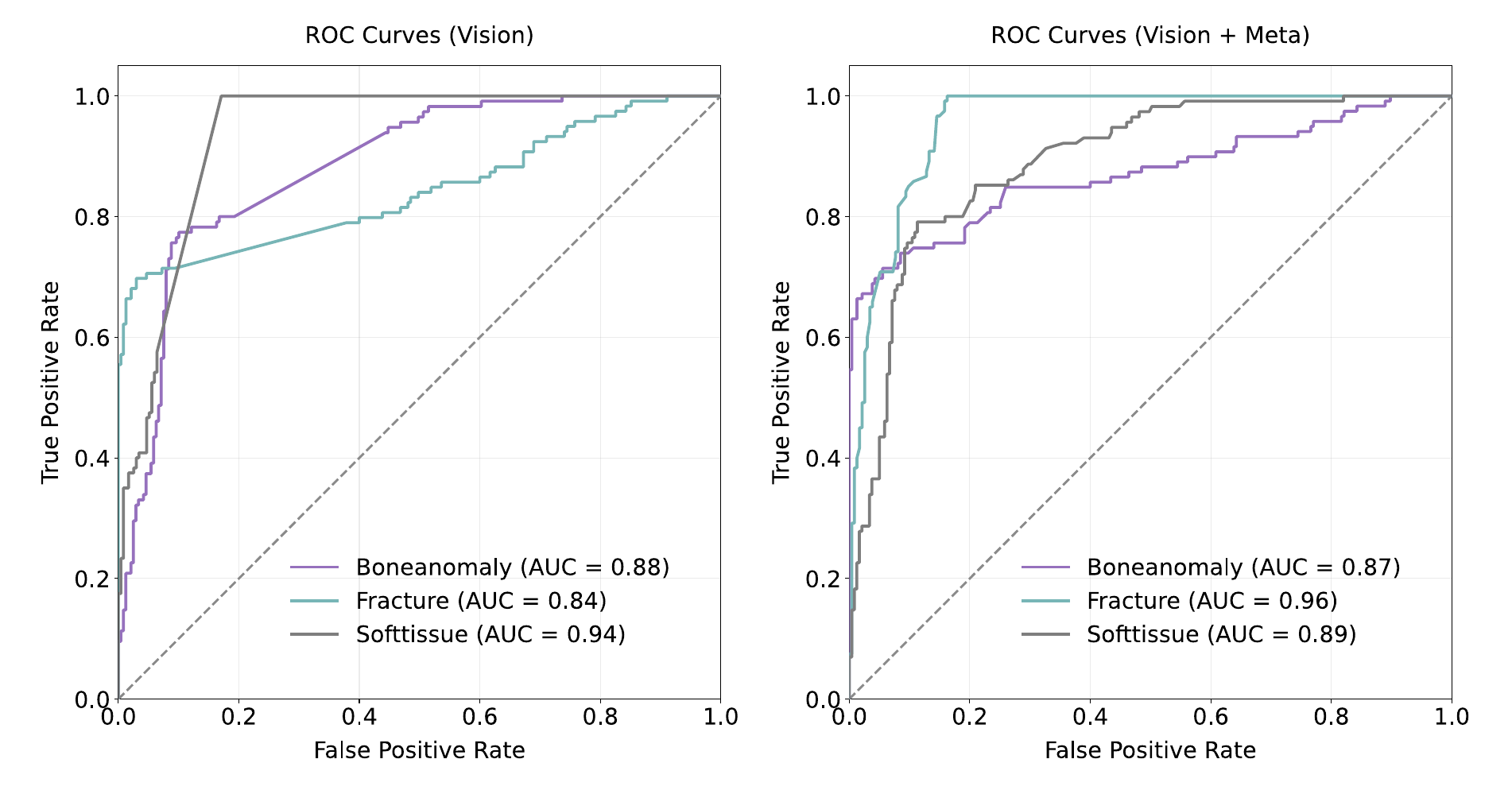}
\caption{ROC curves showing that metadata substantially improves fracture discrimination (AUC 0.84→0.96).}
\label{fig:roc}
\end{figure}

   Fig.\ref{fig:roc} further shows that metadata improves fracture discrimination while maintaining performance on bone anomaly and soft-tissue recognition. On the larger \enquote{Fracture vs. No Fracture} dataset (Table~\ref{tab:fracture_dataset_results}), metadata integration produced a 10\% absolute gain. This gain came from being able to better distinguish ``non-fracture'' instances from the ``fractured'' ones than the baseline. 
   These results suggest that metadata not only aids fine-grained wrist pathology recognition but also enhances scalability to larger clinical datasets.

\subsection{Failure Cases \& Error Analysis}
On the held-out test set ($n=354$), the most common confusions are \texttt{boneanomaly} (0)$\rightarrow$\texttt{softtissue} (2) (25 cases) and \texttt{soft\_tissue} (2)$\rightarrow$\texttt{fracture} (1) (23 cases). Consistent with Table~\ref{tab:sens_spec_prec}, class 0 shows high precision ($\approx96\%$) but lower sensitivity ($\approx62$--$66\%$), indicating conservative \texttt{boneanomaly} predictions. Class 1 attains 100\% sensitivity but 75\% precision, i.e., fractures are rarely missed but some non-fractures are overcalled. Accuracy decreases in older age bins ($\approx0.75$--$0.77$ for $\texttt{age\_norm}\geq0.526$ vs.\ $\approx0.91$ for $\texttt{age\_norm}<0.526$), while performance is similar across sex. Notably, some correct predictions have low confidence (median $\approx0.74$), suggesting borderline cases where clinician review may help.

\begin{table}
\centering
\caption{Performance evaluation on a significantly larger (fracture vs. no fracture) dataset.}
\setlength{\tabcolsep}{10pt}
\renewcommand{\arraystretch}{1.15}
\resizebox{0.5\columnwidth}{!}{%
    \begin{tabular}{l c}
    \hline
    \textbf{FG-2-inat (Ours)} &\textbf{ Test Accuracy (\%)} \\
    \hline 
        Vision & $50.1$\\
        Vision + Meta & $\mathbf{60.4}$\\
    \hline                                                      
  \end{tabular}}
  \label{tab:fracture_dataset_results}
\end{table}

\section{\textbf{Conclusion}}
% In this study, we demonstrated that integrating patient demographic metadata with image data enhances fine-grained recognition of pediatric wrist pathologies. 

Overall, our experiments demonstrate three key findings:  
(1) Fine-grained transformer architecture outperforms conventional and modern CNNs for pediatric wrist pathology.  
(2) Demographic metadata (age, sex) improves recognition across both limited and large-scale datasets. (3) Fine-grained pretraining enhances generalization compared to using ImageNet weights. These results emphasize that integrating developmental context into vision models is essential for robust pediatric musculoskeletal diagnosis.

\begin{table}
    \centering
    \caption{Evaluation of our default fusion (FG-2-inat).}
    \setlength{\tabcolsep}{10pt}
\renewcommand{\arraystretch}{1.15}
    \resizebox{0.8\columnwidth}{!}{%
    \begin{tabular}{ccccc}
        \hline
        \textbf{Configuration} & \textbf{Cls} & \textbf{Sensitivity (\%)} & \textbf{Specificity (\%)} & \textbf{Precision (\%)}\\
        \hline
        \multirow{4}{*}{Vision} & $0$ & $62.2$ & $98.7$ & $96.1$\\
        & $1$ & $100.0$ & $82.9$ & $75.0$ \\
        & $2$ & $77.4$ & $88.3$ & $76.1$\\
        \hline
        \multirow{4}{*}{Vision+Meta} & $0$ & ${66.4}$ & $98.7$ & ${96.3}$\\
        & $1$ & $100.0$ & $82.9$ & $75.0$ \\
        & $2$ & $77.4$ & ${90.4}$ & ${79.5}$\\
        \hline
    \end{tabular}}
    \label{tab:sens_spec_prec}
\end{table}

\section{\textbf{Acknowledgement}}
% This work was supported by the Curricula Development and Capacity Building in Applied Computer Science for Pakistani Higher Education Institutions (CONNECT) Project NORPART-2021/10502, funded by the Norwegian Directorate for Higher Education and Skills (DIKU).
This work was supported by the CONNECT Project NORPART-2021/10502, funded by the Norwegian Directorate for Higher Education and Skills (DIKU).

\bibliographystyle{unsrt}
\bibliography{bibfile}

@article{florian2024, title={Impact of metadata in multimodal classification of bone tumours}, volume={25}, url={https://bmcmusculoskeletdisord.biomedcentral.com/articles/10.1186/s12891-024-07934-9}, DOI={https://doi.org/10.1186/s12891-024-07934-9}, number={1}, journal={BMC Musculoskeletal Disorders}, publisher={BioMed Central}, author={Florian Hinterwimmer and Guenther, Michael and Consalvo, Sarah and Neumann, Jan and Gersing, Alexandra and Klaus Woertler and Rüdiger von Eisenhart-Rothe and Rainer Burgkart and Rueckert, Daniel}, year={2024}, month={Oct} }

@article{adams_robert_yi_babyn_2020, title={Artificial Intelligence Solutions for Analysis of X-ray Images}, volume={72}, url={https://journals.sagepub.com/doi/abs/10.1177/0846537120941671}, DOI={https://doi.org/10.1177/0846537120941671}, number={1}, journal={Canadian Association of Radiologists Journal}, publisher={SAGE Publishing}, author={Adams, Scott J and Robert and Yi, Xin and Babyn, Paul}, year={2020}, month={Aug}, pages={60–72} }

@misc{woo2023convnextv2,
  title         = {ConvNeXt V2: Co-designing and Scaling ConvNets with Masked Autoencoders},
  author        = {Woo, Sanghyun and Debnath, Shoubhik and Hu, Ronghang and Chen, Xinlei and Liu, Zhuang and Kweon, In So and Xie, Saining},
  year          = {2023},
  eprint        = {2301.00808},
  archivePrefix = {arXiv},
  primaryClass  = {cs.CV},
  doi           = {10.48550/arXiv.2301.00808},
  url           = {https://arxiv.org/abs/2301.00808}
}

@article{bian_hand_wrist_bone_age_2020,
  author    = {Bian, Zhen and Guo, Yuan and Lyu, XueMin and Yang, Zheng and Cheung, Jason Pui Yin},
  title     = {Relationship between hand and wrist bone age assessment methods},
  journal   = {Medicine},
  year      = {2020},
  volume    = {99},
  number    = {39},
  pages     = {e22392},
  month     = sep,
  doi       = {10.1097/md.0000000000022392},
  url       = {https://pmc.ncbi.nlm.nih.gov/articles/PMC7523768/},
  publisher = {Wolters Kluwer Health},
}

@article{kox_amphys_protocol_2018,
  author    = {Kox, Laura S. and Kraan, Rik B. J. and van Dijke, Kees F. and Hemke, Robert and Jens, Sjoerd and de Jonge, Milko C. and Oei, Edwin H. G. and Smithuis, Frank F. and Terra, Maaike P. and Maas, Mario},
  title     = {Systematic assessment of the growth plates of the wrist in young gymnasts: development and validation of the Amsterdam {MRI} assessment of the Physis ({AMPHYS}) protocol},
  journal   = {BMJ Open Sport \& Exercise Medicine},
  year      = {2018},
  volume    = {4},
  number    = {1},
  pages     = {e000352},
  month     = apr,
  doi       = {10.1136/bmjsem-2018-000352},
  url       = {https://pmc.ncbi.nlm.nih.gov/articles/PMC5905740/},
  publisher = {BMJ},
}

@article{zhao_fine_grained_segmentation_2017,
  author    = {Zhao, Bo and Feng, Jiashi and Wu, Xiaogang and Yan, Shuicheng},
  title     = {A survey on deep learning-based fine-grained object classification and semantic segmentation},
  journal   = {International Journal of Automation and Computing},
  year      = {2017},
  volume    = {14},
  number    = {2},
  pages     = {119--135},
  doi       = {10.1007/s11633-017-1053-3},
  publisher = {Springer},
}

@article{ahmed2024learning,
  title        = {Learning from the few: Fine‑grained approach to pediatric wrist pathology recognition on a limited dataset},
  author       = {Ahmed, Ammar and Imran, Ali Shariq and Kastrati, Zenun and Daudpota, Sher Muhammad and Ullah, Mohib and Noor, Waheed},
  journal      = {Computers in Biology and Medicine},
  volume       = {181},
  pages        = {109044},
  year         = {2024},
  doi          = {10.1016/j.compbiomed.2024.109044},
  url          = {https://www.sciencedirect.com/science/article/pii/S0010482524011296}
}

@incollection{krizhevsky2012imagenet,  author       = {Krizhevsky, Alex and Sutskever, Ilya and Hinton, Geoffrey E.},  title        = {{ImageNet Classification with Deep Convolutional Neural Networks}},  booktitle    = {Advances in Neural Information Processing Systems},  year         = {2012},  url          = {https://papers.nips.cc/paper/2012/hash/c399862d3b9d6b76c8436e924a68c45b-Abstract.html}}

@article{tan2021efficientnetv2,  author  = {Tan, Mingxing and Le, Quoc V.},  title   = {{EfficientNetV2: Smaller Models and Faster Training}},  year    = {2021},  url     = {https://arxiv.org/abs/2104.00298},  eprint  = {2104.00298},  archivePrefix = {arXiv}}

@article{huang2018densely,  author  = {Huang, Gao and Liu, Zhuang and van der Maaten, Laurens and Weinberger, Kilian Q.},  title   = {{Densely Connected Convolutional Networks}},  year    = {2018},  url     = {https://arxiv.org/abs/1608.06993},  eprint  = {1608.06993},  archivePrefix = {arXiv}}

@article{he2015deep,  author  = {He, Kaiming and Zhang, Xiangyu and Ren, Shaoqing and Sun, Jian},  title   = {{Deep Residual Learning for Image Recognition}},  year    = {2015},  url     = {https://arxiv.org/abs/1512.03385},  eprint  = {1512.03385},  archivePrefix = {arXiv}}

@inproceedings{deng2009imagenet,
  title        = {ImageNet: A Large-Scale Hierarchical Image Database},
  author       = {Deng, Jia and Dong, Wei and Socher, Richard and Li, Li-Jia and Li, Kai and Fei-Fei, Li},
  booktitle    = {Proceedings of the IEEE Conference on Computer Vision and Pattern Recognition (CVPR)},
  pages        = {248--255},
  year         = {2009},
  organization = {IEEE},
  doi          = {10.1109/CVPR.2009.5206848},
  url          = {https://ieeexplore.ieee.org/document/5206848}
}

@article{vanhorn2018inaturalist,
  author  = {Van Horn, Grant and Mac Aodha, Oisin and Song, Yang and Cui, Yin and Sun, Chen and Shepard, Alex and Adam, Hartwig and Perona, Pietro and Belongie, Serge},
  title   = {{The iNaturalist Species Classification and Detection Dataset}},
  journal = {arXiv.org},
  year    = {2018},
  month   = {April},
  day     = {10},
  eprint  = {1707.06642},
}

@article{nagy2022pediatric,
  author    = {Nagy, Eveline and Janisch, Martin and Hr{\v{z}}i{\'c}, Filip and Aresta, Guilherme and Bischof, Horst and others},
  title     = {A pediatric wrist trauma X-ray dataset (GRAZPEDWRI-DX) for machine learning},
  journal   = {Scientific Data},
  year      = {2022},
  volume    = {9},
  pages     = {222},
  doi       = {10.1038/s41597-022-01328-z},
  publisher = {Springer Nature},
}

@article{diao2022metaformer,
  author  = {Diao, Qishuai and Jiang, Yi and Wen, Bin and Sun, Jia and Yuan, Zehuan},
  title   = {{MetaFormer: A Unified Meta Framework for Fine-Grained Recognition}},
  journal = {arXiv.org},
  year    = {2022},
  day     = {5},
  eprint  = {2203.02751},
}

@article{liu2021swintransformer,
  author  = {Liu, Ze and Lin, Yutong and Cao, Yue and Hu, Han and Wei, Yixuan and Zhang, Zheng and Lin, Stephen and Guo, Baining},
  title   = {{Swin Transformer: Hierarchical Vision Transformer using Shifted Windows}},
  journal = {arXiv preprint},
  eprint  = {2103.14030},
  year    = {2021},
  month   = {3},
  day     = {4},
  note    = {arXiv:2103.14030}
}

@article{dosovitskiy2020image,
  author  = {Dosovitskiy, Alexey and Beyer, Lucas and Kolesnikov, Alexander and Weissenborn, Dirk and Zhai, Xiaohua and Unterthiner, Thomas and Dehghani, Mostafa and Minderer, Matthias and Heigold, Georg and Gelly, Sylvain and Uszkoreit, Jakob and Houlsby, Neil},
  title   = {{An Image is Worth 16x16 Words: Transformers for Image Recognition at Scale}},
  year    = {2020},
  eprint  = {2010.11929},
  archivePrefix = {arXiv}
}

@inproceedings{szegedy2015going,
  author      = {Szegedy, Christian and Liu, Wei and Jia, Yangqing and Sermanet, Pierre and Reed, Scott and Anguelov, Dragomir and Erhan, Dumitru and Vanhoucke, Vincent and Rabinovich, Andrew},
  title       = {{Going deeper with convolutions}},
  booktitle   = {2015 IEEE Conference on Computer Vision and Pattern Recognition (CVPR)},
  year        = {2015},
  organization= {IEEE},
  pages       = {1--9},
}

@article{ningrum2021deep,
  author       = {Ningrum, Dina Nur Anggraini and Yuan, Sheng-Po and Kung, Woon-Man and Wu, Chieh-Chen and Tzeng, I-Shiang and Huang, Chu-Ya and Li, Jack Yu-Chuan and Wang, Yao-Chin},
  title        = {{Deep Learning Classifier with Patient’s Metadata of Dermoscopic Images in Malignant Melanoma Detection}},
  journal      = {Journal of Multidisciplinary Healthcare},
  volume       = {14},
  year         = {2021},
  pages        = {877--885},
  doi          = {10.2147/JMDH.S306284},
}

@incollection{nunnari2020study,
  author       = {Nunnari, Fabrizio and Bhuvaneshwara, Chirag and Ezema, Abraham Obinwanne and Sonntag, Daniel},
  title        = {{A Study on the Fusion of Pixels and Patient Metadata in CNN-Based Classification of Skin Lesion Images}},
  booktitle    = {Machine Learning and Knowledge Extraction},
  editor       = {Holzinger, Andreas and Kieseberg, Peter and Tjoa, A Min and Weippl, Edgar},
  publisher    = {Springer International Publishing},
  year         = {2020},
  pages        = {191--208},
  doi          = {10.1007/978-3-030-57321-8_11},
}

@article{cai2023multimodal,
  author       = {Cai, Gan and Zhu, Yu and Wu, Yue and Jiang, Xiaoben and Ye, Jiongyao and Yang, Dawei},
  title        = {{A Multimodal Transformer to Fuse Images and Metadata for Skin Disease Classification}},
  journal      = {The Visual Computer},
  volume       = {39},
  number       = {7},
  year         = {2023},
  pages        = {2781--2793},
  doi          = {10.1007/s00371-022-02492-4},
}

@inproceedings{thomas2021combining,
  author       = {Thomas, Spencer A.},
  title        = {{Combining Image Features and Patient Metadata to Enhance Transfer Learning}},
  booktitle    = {2021 43rd Annual International Conference of the IEEE Engineering in Medicine and Biology Society (EMBC)},
  year         = {2021},
  pages        = {2660--2663},
  doi          = {10.1109/EMBC46164.2021.9630047},
}

@misc{hoynak2022wrist,
  author       = {Hoynak, Bryan C. MD},
  title        = {{Wrist Fracture Management in the ED}},
  howpublished = {Practice Essentials, Pathophysiology, Prognosis},
  organization = {Medscape},
  year         = {2022},
}

@article{roche_skeletal_maturity_1974,
  author      = {Roche, Alex F. and Roberts, Jean and Hamill, Peter V. V.},
  title       = {Skeletal maturity of children 6--11 years, United States},
  institution = {National Center for Health Statistics},
  year        = {1974},
  month       = nov,
  series      = {Vital and Health Statistics. Series 11, Data from the National Health Survey},
  number      = {140},
  pages       = {1--62},
  url         = {https://pubmed.ncbi.nlm.nih.gov/4374802/},
  note        = {PMID: 4374802},
}

@misc{cui_large_scale_fine_grained_2018,
  author       = {Cui, Yin and Song, Yang and Sun, Chen and Howard, Andrew and Belongie, Serge},
  title        = {Large Scale Fine-Grained Categorization and Domain-Specific Transfer Learning},
  year         = {2018},
  eprint       = {1806.06193},
  archivePrefix= {arXiv},
  primaryClass = {cs.CV},
  url          = {https://arxiv.org/abs/1806.06193},
}

@software{Jocher_Ultralytics_YOLO_2023,
author = {Jocher, Glenn and Qiu, Jing and Chaurasia, Ayush},
license = {AGPL-3.0},
month = jan,
title = {{Ultralytics YOLO}},
url = {https://github.com/ultralytics/ultralytics},
version = {8.0.0},
year = {2023}
}
\end{document}